 \documentclass[final,3p,times,twocolumn]{elsarticle}
\usepackage{amssymb}
\usepackage{hyperref}
\usepackage{graphicx}
\usepackage{subcaption}
\usepackage{lscape}
\usepackage{amsmath}
\usepackage{multirow}
\usepackage[figuresright]{rotating}
\usepackage{lineno}
\usepackage{algorithm} 
\usepackage{algpseudocode} 
\usepackage{comment}
\usepackage{lineno}
\usepackage{hyperref}
\usepackage{booktabs} % for improved table lines (optional)
\usepackage{rotating} % provides sidewaystable and sidewaysfigure

\usepackage{natbib}
\setcitestyle{authoryear,open={(},close={)},citesep={;}} 

\usepackage{calc}
\begin{document}

\begin{frontmatter}

\title{Semi-Supervised Weed Detection in Vegetable Fields: In-domain and Cross-domain Experiments}

\author[label1]{Boyang Deng}
\author[label1]{Yuzhen Lu}
\author[label2]{Daniel Brainard}

\address{Yuzhen Lu (luyuzhen@msu.edu) is the corresponding author}
\address[label1]{Department of Agricultural and Biosystems Engineering, Michigan State University, East Lansing, MI 48824, USA}
\address[label2]{Department of Horticulture, Michigan State University, East Lansing, MI 48824, USA}

\begin{abstract}
Robust weed detection remains a challenging task in precision weeding, requiring not only potent weed detection models but also large-scale, labeled data. However, the labeled data adequate for model training is practically difficult to come by due to the time-consuming, labor-intensive process that requires specialized expertise to recognize plant species. This study introduces semi-supervised object detection (SSOD) methods for leveraging unlabeled data for enhanced weed detection and proposes a new YOLOv8-based SSOD method, i.e., WeedTeacher. An experimental comparison of four SSOD methods, including three existing frameworks (i.e., DenseTeacher, EfficientTeacher, and SmallTeacher) and WeedTeacher, alongside fully supervised baselines, was conducted for weed detection in both in-domain and cross-domain contexts. A new, diverse weed dataset was created as the testbed, comprising a total of 19,931 field images from two differing domains, including 8,435 labeled (basic-domain) images acquired by hand-holding devices from 2021 to 2023 and 11,496 unlabeled (new-domain) images acquired by a ground-based mobile platform in 2024. The in-domain experiment with models trained using 10\% of the labeled, basic-domain images and tested on the remaining 90\% of the data, showed that the YOLOv8-based WeedTeacher achieved the highest accuracy among all four SSOD methods, with an improvement of 2.6\% mAP@50 and 3.1\% mAP@50:95 over its supervised baseline (i.e., YOLOv8l). In the cross-domain experiment where the unlabeled new-domain data was incorporated, all four SSOD methods, however, resulted in no or limited improvements over their supervised counterparts. Research is needed to address the difficulty of cross-domain data utilization for robust weed detection.

\end{abstract}

\begin{keyword}
Domain shift, Precision agriculture, Deep semi-supervised learning, Weed detection.
\end{keyword}

\end{frontmatter}

%%%%%%%%%%%%%%%%%%%%%%%%%%%%%%%%%%%%%%%%%%%%%%%%%%%%%%%%%%%%%%%%%%%%%%%%%%%%%%%%%%%%%%%%%%%%%%%%%%%%%%%%%%%%%%%%%%%%%%%%%%%%%%%%%%%%%%
%% main text
\section{Introduction}
\label{sec:intro}

Weeds pose a persistent challenge in crop production, as the competition between weeds and crops for resources (e.g., light, nutrients, and water) can dramatically diminish crop yield and quality. Among various biotic stressors for crops, weeds account for an average of 31.5\% of global yield losses, resulting in an annual economic loss of over \$30 billion  \citep{Kubiak2022}. Weed management in vegetable crops faces particular challenges due to the vulnerability of the cropping systems \citep{westwood2018weed} and the limited herbicide options available \citep{Boyd2022}. Vegetables often have a shorter growing season and less competitive ability, making them susceptible to yield losses when weeds are not well controlled \citep{Pannacci2017}. Significant efforts including manual labor are thus often needed to effectively manage weeds in vegetables.

Machine vision technologies have advanced rapidly in recent years, becoming a powerful tool for weed recognition and precision weed control \citep{li2022}. Taking advantage of imaging sensors and dedicated machine vision algorithms for plant perception, site-specific weed management promises to be practically implemented. Significant progress has been made in developing various machine vision-based weeding systems \citep{Machleb2020, Allmendinger2022, Merfield2023}. For example, \citet{bawden2017} designed the AgBotII robotic prototype that integrated with vision-based plant detection and classification for targeted treatment of individual weed species. Their system achieved 92.3\% accuracy in identifying four weed species in a cotton field. The weeding actuator consisted of a hoe-style mechanical module and a sprayer for weed-species-specific treatment based on weed detection, although incorrect treatments occurred due to misclassified weed species \citep{bawden2017}. Recently, \citet{Upadhyay2024} developed a YOLOv4-based smart sprayer system to achieve an average effective spraying rate of 90.6\% in their field experiment. However, their field experiment was conducted on a single weed species, and the error rate of 9.4\% was due to inaccurate weed detection. Therefore, although significant progress has been made, improving weed detection performance remains a demanding task for practical weed management.

As artificial intelligence (AI) is impacting agriculture, deep learning (DL) has become the cutting-edge for agricultural applications, beyond traditional image processing and feature-engineering approaches. However, DL architectures generally require large-scale, annotated datasets for supervised training, which is crucial for developing robust machine vision systems capable of adapting to diverse field conditions. Over the past ten years, numerous weed datasets have been created and made publicly available \citep{Lu2020, Deng2024Weed}, although most of these datasets may be insufficient to ensure satisfactory results because of limited representations of natural variability in weeds/crops and environmental factors. Compared to data collection processes, annotating acquired images is indeed a rate-limiting bottleneck due to the resources and expertise needed to identify plant species and label individual instances, rendering it difficult and costly to label large volumes of data. In this regard, there is growing interest in pursuing semi-supervised learning (SSL) techniques \citep{Yang2022} to boost the performance of supervised models while saving efforts for image annotation, by exploiting extensive amounts of unlabeled data and jointly learning from labeled and unlabeled data.

As an extension of SSL traditionally for image classification, semi-supervised object detection (SSOD) focuses on bounding box-based object detection tasks \citep{wang2023}, which generally involves implementing a teacher-student learning framework, i.e., firstly training an initial detector (teacher) using labeled data, then generating pseudo labels by the trained detector (teacher) and boxes on unlabeled data, and finally utilizing them alongside the existing labeled data to retrain the detection model (student). Different SSOD methods have been proposed over the past five years in the computer vision field, which differ mainly in pseudo-label selection and modeling architectures. For instance, DenseTeacher \citep{Zhou2022} introduced dense pseudo-maps and a region division strategy selection technique to enhance training without complex post-processing, while EfficientTeacher \citep{Xu2023} aimed to improve pseudo-label consistency and training efficiency for one-stage, anchor-based detector. It is noted that most existing SSOD techniques were not designed for or may not generalize well to agricultural images. 

Currently, there are only a handful of studies on SSL-based weed recognition. \citet{Tseng2023} proposed SmallTeacher with calibrated teacher-student learning and FasterRCNN for weed detection, achieving a mAP@50 improvement of 3.88\% on a two-class dataset with 960 labeled images, compared to fully supervised baselines. \citet{Li2024} reported improved detection through SSL on two public cotton weed datasets \citep{dang2023, Rahman2023}. To achieve a better balance between accuracy and efficiency, \citet{Saleh2024} adapted the one-stage detector YOLOv5 for semi-supervised weed detection, where multi-scale feature extraction was implemented in the detector and a dynamic pseudo-label assigner was used with a modified overall training loss to reduce the effect of unreliable labels. All these studies did not examine cross-domain weed detection, which is important for real-world applications.

Inspired by SmallTeacher \citep{Tseng2023} and MixPL \citep{Chen2023}, this study proposes a new WeedTeacher that integrates YOLOv8 for enhanced weed detection. The main contributions of this work are three-fold: 1) the first integration of SSOD with YOLOv8 for multi-class weed detection, 2) a new, diverse dataset from four seasons with a total of 19,931 field images as the weed detection testbed, and 3) performance comparison of WeedTeacher and three existing SSOD methods (i.e., EfficientTeacher, DenseTeacher, SmallTeacher), alongside their supervised baselines, for weed detection in both in-domain and cross-domain contexts.

% You may continue the edit from here 
\section{Materials and Methods}
\label{sec:data_collection}
%%%%%%%%%%%%%%%%%%%%%%%%%%%%%%%%%%%%%%%%%%%%%%%%%%%%%%%%%%%%%%
\subsection{Dataset Preparation}

The weed dataset here consists of two subsets representing two distinct domains. One subset is the public 3SeasonWeedDet10 dataset \citep{Lu2025}, which consists of 8,436 images of 10 weed classes; it was created in our earlier study to assess stable diffusion for weed image generation \citep{Deng2025}. The images were collected over three consecutive years, including 4,704 in 2021, 1,948 in 2022, and 1,784 in 2023, under natural light conditions from diverse field sites in Mississippi and Michigan, and most of them were captured using handheld color cameras or smartphones. This dataset was fully labeled with 27,963 bounding boxes for weed instances. A detailed description of data curation and statistics is given in \citet{Lu2025} and \citet{Deng2025}. 

\begin{figure}[!ht]
 \centering
 \includegraphics[width=\columnwidth]{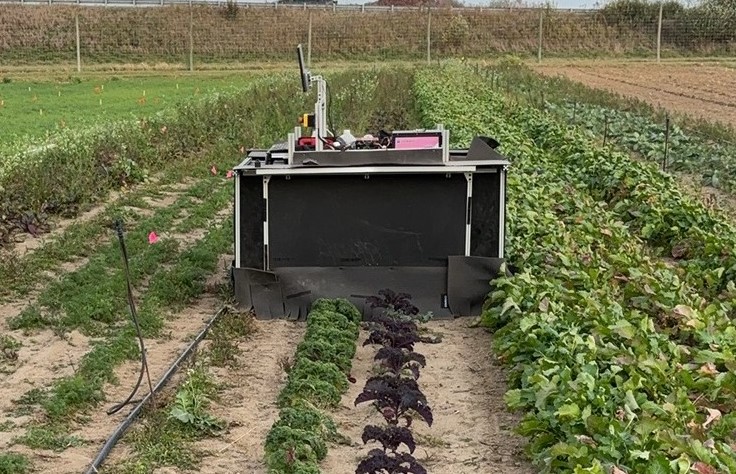}
 \caption{The ground-based mobile platform for acquiring images in vegetable fields in 2024.}
 \label{fig:platform}
\end{figure}

The other subset contains 11,496 unlabeled images acquired by a ground-based mobile platform \citep{Deng2024Weeding}, as shown in Figure~\ref{fig:platform}. A color camera, with an output image resolution of 1616×2256 pixels (after applying a region of interest in the camera setup), was mounted at a height of 70 cm above the ground, pointing downwards and covering the width of approximately 106 cm (2-3 crop rows). The platform was enclosed by black panels to suppress the effect of variable field light, and light sources were configured to provide controlled lighting during data collection. Images were collected in experimental field plots with five vegetable crops (i.e., turnips, carrots, cabbage, onions, and beets), grown under weed management, on the horticultural farm (Holt, MI) of Michigan State University, during September and October of 2024. This subset shares the primary weeds in the first subset (3SeasonWeedDet10) but with crop plants present in acquired images, and it is unlabeled.

\begin{figure*}[!ht]
 \centering
 \includegraphics[width=\linewidth]{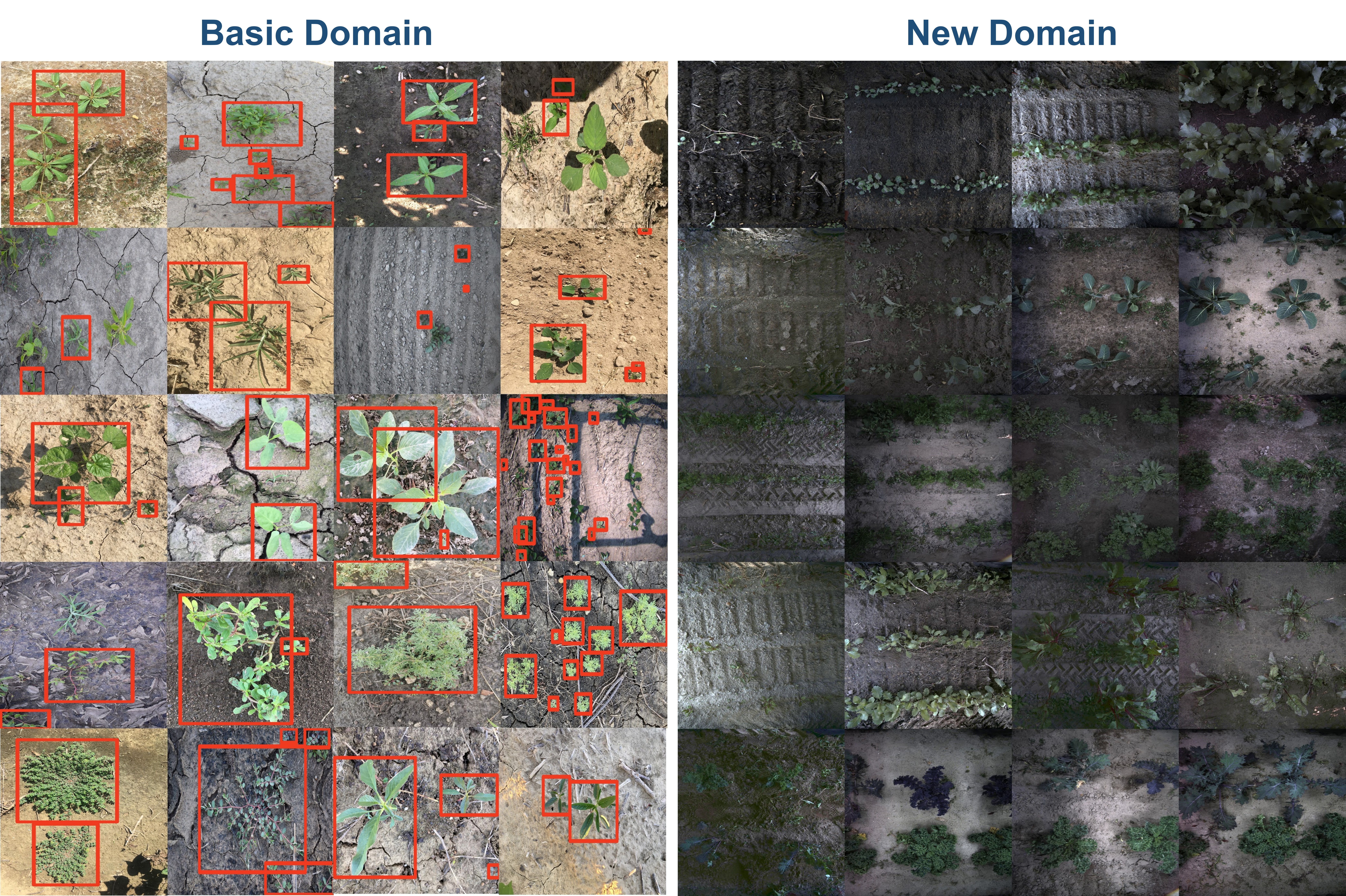}
 \caption{Example images from the labeled basic domain (left) and the unlabeled new domain (right). }
 \label{fig:weeds}
\end{figure*}

Since the images in the two subsets have distinct distributions from vision examination (Figure~\ref{fig:weeds}), because of different methods for data collection and other factors (e.g., field conditions), the two sets were considered to represent different domains. The new domain subset poses added challenges to weed detection, because of comparatively low image quality (e.g., even lighting, complex field scenes) (Figure~\ref{fig:weeds}). In this study, the fully labeled subset is regarded as the basic domain, while the second unlabeled set is referred to as the new domain. 

\subsection{SSOD for In-domain and Cross-domain Weed Detection}
Three existing SSOD frameworks were examined in this study for semi-supervised weed detection, that is, \textbf{DenseTeacher} \citep{Zhou2022}, \textbf{EfficientTeacher} \citep{Xu2023}, and \textbf{SmallTeacher} \citep{Tseng2023}. The first two methods were chosen given their demonstrated performance in generic object detection tasks and the availability of their open-source implementation. DenseTeacher stands out for its focus on dense predictions, extending the teacher’s supervision beyond bounding boxes to finer details across the feature map, improving the detection of small or overlapping objects. EfficientTeacher induces a pseudo-label assigner for a refined use of pseudo labels from dense detectors and an epoch adaptor method for stable and efficient end-to-end training. As done in the original implementation of the two SSOD methods, DenseTeacher and EfficientTeacher were applied to FCOS \citep{Tian2020} and YOLOv5 \citep{Jocher2020}, respectively, for weed detection here.

\begin{figure}[!ht]
 \centering
 \includegraphics[width=\columnwidth]{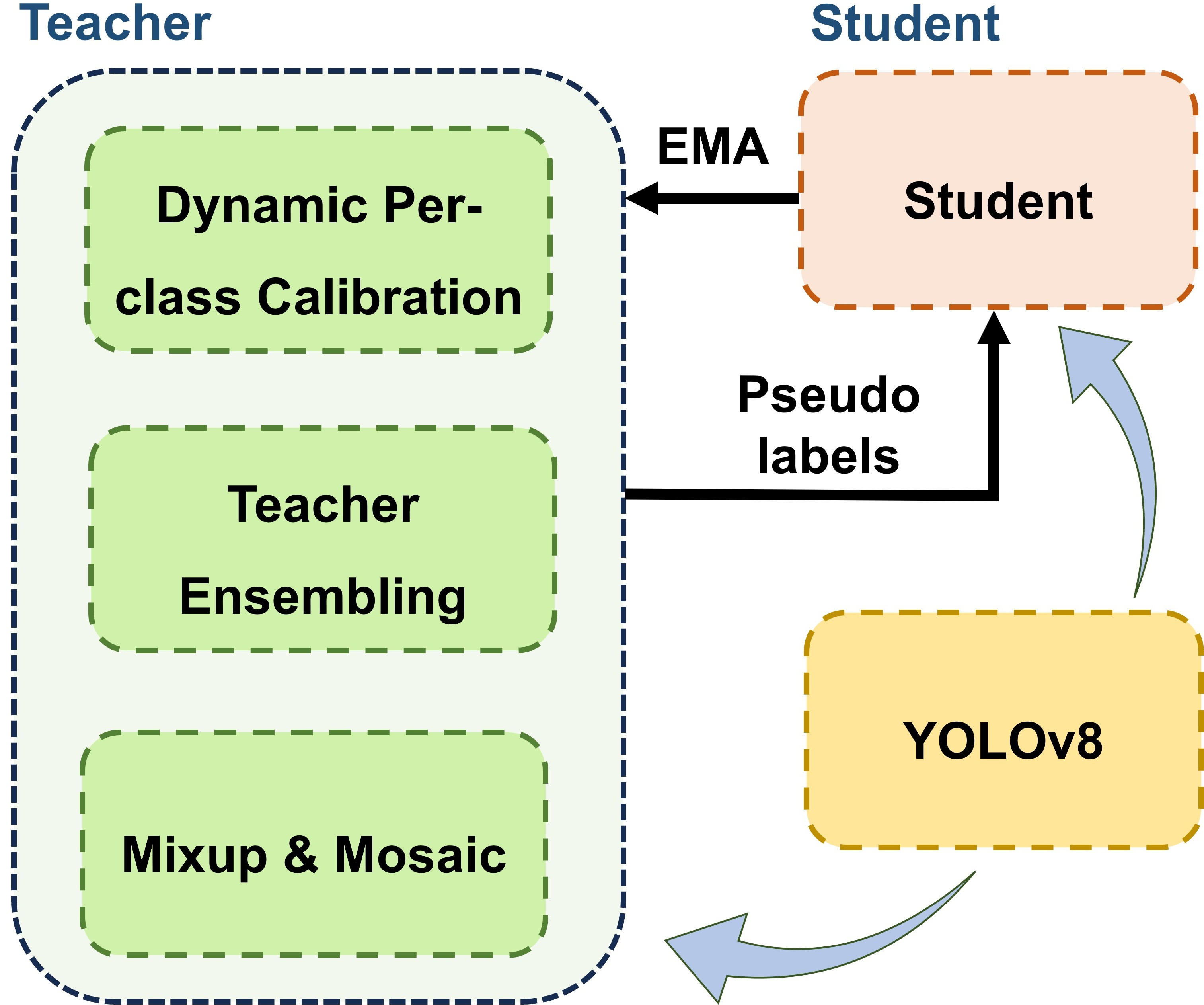}
 \caption{YOLOv8-based WeedTeacher framework. EMA denotes exponential moving average. Teacher and student models share the same YOLOv8-large architecture.}
 \label{fig:WeedTeacher}
\end{figure}

SmallTeacher \citep{Tseng2023} was perhaps the first method proposed specifically for agricultural images, which is featured by calibrated teacher-student learning through dynamic per-class calibration, weak teacher ensembling using test-time augmentations, and weighted boxes fusion (WBF) \citep{Solovyev2021}. The SmallTeacher method was applied to Faster-RCNN \citep{Ren2016} for weed detection on a small, two-class dataset of 960 images with the crops and weeds labeled. Although it achieved improvements over full-supervised baselines, the method did not generalize well to weed datasets involving more complex field scenes and weed classes based on our experimentation. Moreover, fast, source-efficient one-stage object detectors with high accuracy are more desirable for real-world applications. In this regard, a new SSOD method that integrates YOLOv8 \citep{Jocher2023} for weed detection, i.e., WeedTeacher, was proposed in this study. Figure ~\ref{fig:WeedTeacher} shows the framework of the proposed WeedTeacher.

WeedTeacher integrates the dynamic per-class confidence calibration and the WBF-based weak teacher ensembling of SmallTeacher into the YOLOv8 semi-supervised training pipeline. To enhance efficiency, up to 1,000 unlabeled images were utilized for per-epoch threshold updates in WeedTeacher. The YOLOv8-based teacher model initialization and updates employ the YOLOv5-based EfficientTeacher mechanism, which can be seamlessly transferred to the YOLOv8 training process. Inspired by MixPL \citep{Chen2023}, which shows that combining Mixup and Mosaic augmentations helps reduce the negative impact of missed and incorrect pseudo-labels while introducing meaningful perturbations for accurate positive samples, the two advanced augmentations are thus incorporated in WeedTeacher during data augmentation of teacher-predicted unlabeled images in training the student model. In addition, since student models often struggle with predicting unlabeled images at the beginning, the first half of the training epochs serve as a burn-in phase to establish a stable foundation before incorporating pseudo-labels from the teacher model. The WeedTeacher process is applied to the second half of the training process. 

\subsection{Experimental Setup} 
\begin{figure}[!ht]
 \centering
 \includegraphics[width=\columnwidth]{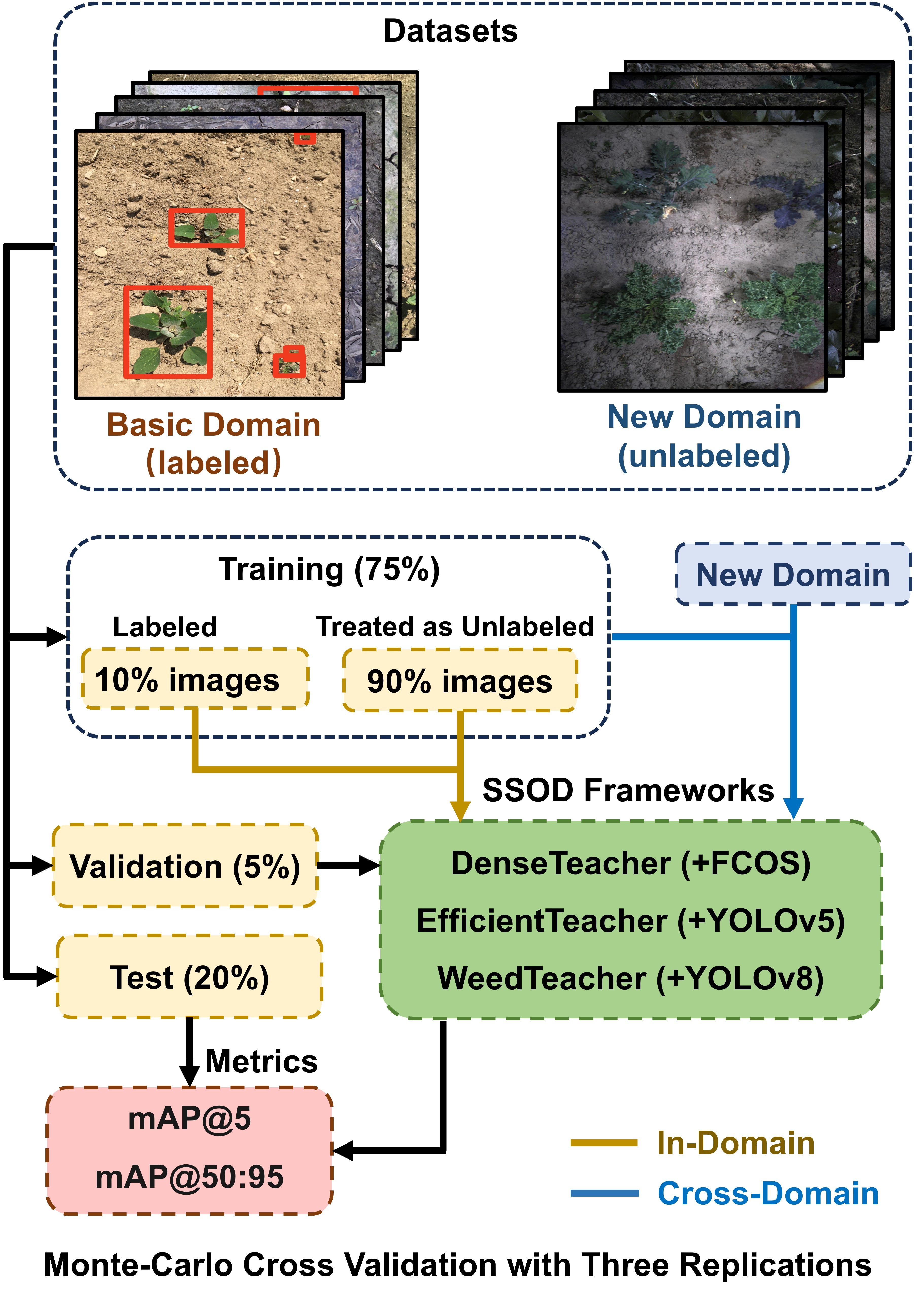}
 \caption{The modeling pipeline of semi-supervised weed detection experiments. The yellow path represents modeling data within the basic domain, while the blue path indicates the experiment incorporating the new domain images as unlabeled data.}
 \label{fig:Pipline}
\end{figure}

Two experiments were conducted to evaluate SSOD methods for weed detection. First, models were trained using 10\% of labeled images from the basic domain, with the remaining 90\% treated as unlabeled, which is referred to as an in-domain weed detection configuration. In the second experiment, all basic domain images were used as labeled, while the unlabeled new domain images were incorporated to train models, which is referred to as a cross-domain detection setup. In each scenario, fully supervised baseline models were trained for comparison. Figure~\ref{fig:Pipline} illustrates the modeling pipeline.

The DenseTeacher and EfficientTeacher models were trained using the open-sourced packages \citep{Zhou2022, Xu2023}, with default hyperparameters, except that the learning rate for the DenseTeacher was empirically adjusted to 0.001. Likewise, for the WeedTeacher, the default hyperparameters of YOLOv8l (large) were applied, including a learning rate of 0.01. Original high-resolution weed images were resized to the size of 960 × 960 pixels by scaling the longer side to 960 and padding the shorter side to maintain the aspect ratio. All models were trained for 36 epochs with a batch size of 4, which allowed adequate convergence from empirical observations. 

Throughout the training process for all the SSOD frameworks,
the exponential moving average (EMA) was employed for
updating teacher models. Moreover, as in our prior study \citep{Deng2024Weed}, the Monte-Carlo cross-validation with three replications, in which the dataset was split into training, validation, and testing sets according to a ratio of 75\% : 5\% : 20\% using a different seed in each replication. Performance metrics on the test set were averaged across the three replications for model assessment. The experiments were conducted using Python packages including PaddlePaddle (v3.0.0) for DenseTeacher and PyTorch (v1.13.1) for SmallTeacher, EfficientTeacher, and WeedTeacher on a workstation with an Intel i9-10900X CPU (256 GB RAM) and NVIDIA RTX A6000 GPU (48 GB).

\subsection{Performance Metrics}
Weed detection performance, in both semi-supervised and fully supervised schemes, was assessed using mean Average Precision (mAP). The metrics included mAP@50, based on an Intersection over Union (IoU) at a threshold of 0.5, and mAP@50:95, which represents the average mAP across IoU thresholds from 0.50 to 0.95 in an increment of 0.05. The two metrics are commonly used for assessing object detectors in various applications. The basic formula for mAP is as follows:

\begin{equation} \label{eqn:mAP}
mAP  = \frac{1}{N}\sum_{i=1}^{N}AP_{i}
\end{equation}

where \textit{N} represents the number of weed classes, i.e., 10 in this study, and $AP_{i}$ denotes the average precision for class \textit{i}.

%%%%%%%%%%
\section{Results and Discussion}
\subsection{In-domain Weed Detection}

Table 1 shows the in-domain weed detection results. Among four supervised baselines trained with labeled data, YOLOv8l gave the best accuracy, followed by YOLOv5l, FCOS, and Faster-RCNN. For the four SSOD frameworks, SmallTeacher gave the lowest accuracy and slightly underperformed its supervised baseline (Faster-RCNN), indicating deficiencies in the framework. DenseTeacher experienced a marked drop of 13.4\% in mAP@50 (-13.4\%) and 12.5\% in mAP@50:95, compared to its YOLOv5 baseline. The poor performance remains to be ascertained, despite the demonstrated efficacy of DenseTeacher on generic computer vision datasets (e.g., COCO) \citep{Zhou2022}. In contrast, EfficientTeacher achieved improvements of 11.9\% in mAP@50 and 7.9\% in mAP@50:90, over its supervised counterparts. Compared to EfficientTeacher, the proposed YOLOv8-based WeedTeacher yielded smaller (2.6\% in mAP@50 and 3.1 in mAP@50:95) yet meaningful improvements, and notably, it outperformed three other SSOD frameworks by a large margin, which might be largely attributed to the use of YOLOv8l in the framework. Although it is tempting to upgrade FCOS and YOLOv5l with YOLOv8l in DenseTeacher and EfficientTeacher, respectively, the difficulty in implementing the two frameworks on such a more advanced detector as YOLOv8, which remains to be addressed, hindered the exploitation of the two frameworks for a fairer performance comparison. 

\begin{table}[htbp]
\centering
\caption{In-domain weed detection on the basic domain dataset (Figure~\ref{fig:Pipline}). “Full” and “Semi” denote the fully supervised and semi-supervised models, respectively. The “Full” model is based on 10\% labeled training data, while the “Semi” additionally incorporates the remaining 90\% of data treated as unlabeled.}
\label{tab:table1}
\resizebox{0.5\textwidth}{!}{%

\begin{tabular}{lcll}
\hline
Frameworks                  &      & mAP@50      & mAP@50:95   \\
\hline
DenseTeacher                & Full & 66.9         & 43.6         \\
(FCOS)                      & Semi & 53.5 (-13.4) & 31.1 (-12.5) \\
EfficientTeacher            & Full & 61.5         & 42.5         \\
(YOLOv5l)                   & Semi & 73.4 (+11.9) & 50.4 (+7.9)  \\
SmallTeacher                & Full & 53.8         & 25.5         \\
(Faster R-CNN)              & Semi & 53.1 (-0.7)  & 24.9 (-0.6)  \\
\textbf{WeedTeacher}        & Full & 85.0         & 74.1         \\
(YOLOv8l)                   & Semi & 87.6 (+2.6)  & 77.2 (+3.1)  \\
\hline
\end{tabular}
}
\end{table}

% Upon further exploration of EfficientTeacher, it was found that the detection accuracy could be further improved by reinitializing the student and teacher models using the EMA (Exponential Moving Average) of the student model in each semi-supervised epoch. However, the reinitialization of EMA strategy did not result in improvements for the WeedTeacher. The exact reason behind the improvement observed with this setup requires further investigation.

\subsection{Cross-domain Weed Detection}
For the cross-domain weed detection (Table~\ref{tab:table2}), the supervised baseline detectors showed a similar trend, with YOLOv8l remaining the best detector. With more labeled data for training, all models improved substantially compared to those in the in-domain experiment where only 10\% of labeled data were included. For SSOD, DenseTeacher slightly improved over its supervised baseline, although the method performed unsatisfactorily in the in-domain experiment. In contrast, all the other SSOD frameworks did not yield improvements but instead experienced a slight degradation. Overall, there was no or limited impact due to incorporating unlabeled data from a new domain on weed detection, underscoring the challenge with cross-domain weed detection. In such contexts, even high-confidence pseudo-labels predicted by the teacher model could be inaccurate, thus compromising student model performance. Research is needed to explore more effective approaches for unlabeled, cross-domain data utilization.   

\begin{table}[htbp]
\centering
\caption{Cross-domain weed detection (Figure~\ref{fig:Pipline}). “Full” uses all basic domain data alone while “Semi” additionally incorporates unlabeled data from a new domain.}
\label{tab:table2}
\resizebox{0.5\textwidth}{!}{%
\begin{tabular}{lcll}
\hline
Frameworks             &      & mAP@50      & mAP@50:95   \\
\hline
DenseTeacher           & Full & 87.9        & 73.8        \\
(FCOS)                 & Semi & 88.5 (+0.6) & 74.2 (+0.4) \\
EfficientTeacher       & Full & 93          & 79.9        \\
(YOLOv5l)              & Semi & 92.6 (-0.4) & 80.8 (+0.9) \\
SmallTeacher           & Full & 82.8        & 61.9        \\
(Faster R-CNN)         & Semi & 82.7 (-0.1) & 61.8 (-0.1) \\
\textbf{WeedTeacher}   & Full & 94.6        & 87          \\
(YOLOv8l)              & Semi & 94.5 (-0.1) & 86.7 (-0.3) \\
\hline
\end{tabular}
}
\end{table}

\subsection{Discussion}
The utilization of unlabeled, in-domain images for SSOD has been explored in many studies \citep{Zhou2022, Xu2023, Tseng2023}. In this context, the common practice to evaluate SSOD performance is to split a fully labeled dataset into various proportions, designating a subset as labeled and the rest as unlabeled, followed by validation on different portions of labeled data. In this study, incorporating unlabeled new-domain images was evaluated for the first time. As shown in Table 2, using unlabeled label images from a new, differing domain, for even the state-of-the-art SSOD methods, results in minimal or no improvement compared to supervised baselines, indicating the challenges with cross-domain weed detection \citep{Deng2024Weed}. The approaches specifically for cross-domain object detection, such as HarmoniousTeacher (Deng et al., 2023) and VersatileTeacher \citep{Yang2025}, and domain adaptation techniques can potentially help. An intuitive strategy may involve annotating a small portion of the new domain to enable the adaptation of models to the new domain \citep{Inoue2018}. These approaches need further exploration in future work.

The rapid advancements of object detectors demand SSOD frameworks that are easily adaptable. Given the diversity of two-stage anchor-based, one-stage anchor-free, and one-stage anchor-based detectors, it is desirable for SSOD frameworks that can readily adapt to different types of detectors instead of being tied to specific detector architectures. Although it is difficult to predict which type of detector will achieve state-of-the-art performance at any given time, generally newer versions of the detectors tend to improve over older ones, and the performance gain due to implementing more advanced detectors often outweighs the benefits of the SSOD framework itself. Therefore, scaling the SSOD to advanced detectors can be more rewarding than optimizing the framework solely for older detectors.

Currently, most existing SSOD frameworks are designed for or evaluated on older detection architectures (e.g., Faster R-CNN, FCOS, and YOLOv5), and adapting them to newer, more advanced detectors is not a trivial task. The complexity is caused by the intricate implementation requirements of SSOD, such as integrating the teacher model into the detection pipeline and modifying student losses or introducing new loss functions that involve the teacher model. Modifications to loss functions create difficulties when implementing SSOD in new detectors with innovative structures. Hence, achieving SSOD by primarily relying on pre-epoch or teacher-based operations (e.g., dynamic per-class calibration and weak teacher ensembling) along with effective data augmentation can greatly enhance the transferability of the SSOD framework. The WeedTeacher proposed in this study, which requires no modifications to the detector’s loss functions, offers flexibility and compatibility with evolving detector designs and thus the ease of integration with advanced detectors such as YOLOv8. The integration with the latest versions of YOLO detectors, such as YOLOv11 \citep{Jocher2024} and YOLOv12 \citep{Tian2025}, will be pursued in future work.

%%%%%%%%%%%%%%%%%%%%%%%%%%%%%%%%%%%%%%%%%%%%%%%%%%%%%%%%%%%%%%%%%%%%%%%%%%%%%%%%%%%%%%
\section{Conclusion}
\label{sec:conclu}
This study presents the new WeedTeacher that integrates with YOLOv8l for enhanced weed detection through leveraging unlabeled data, and a comparison of the method with three existing SSOD methods (i.e., DenseTeacher, EfficientTeacher, and SmallTeacher) alongside supervised baselines for both in-domain and cross-domain weed detection. A new four-season weed dataset consisting of 19,931 field images, with 11,496 unlabeled from a new domain, was created as the testbed. For in-domain detection, WeedTeacher performed the best among all four SSOD methods and improved over the YOLOv8l baseline by 2.6\% mAP@50 and 3.1\% mAP@50:95. The cross-domain experiment resulted in no or minimal efficacy of all four SSOD methods for weed detection, compared to their supervised baselines, highlighting the challenges of utilizing unlabeled cross-domain images. Research is needed to develop approaches for enhanced cross-domain weed detection while effectively leveraging unlabeled new-domain data.

\section*{Acknowledgment}
This work was supported by the Discretionary Funding Initiative of Michigan State University and the Farm Innovation Grant of the Michigan Department of Agriculture and Rural Development.

\typeout{}
\bibliography{SSOD_ref}
\end{document}